\documentclass{iopconfser}
\usepackage{comment}
\usepackage{svg}
\usepackage{graphicx}      
\usepackage{amsmath}
\usepackage{amssymb}
\usepackage{algorithm}
\usepackage{algorithmic}
\usepackage{caption}
\usepackage{subcaption}
\usepackage[title]{appendix}
\usepackage{multirow}
\usepackage{commath}
\usepackage{mathtools}
\usepackage{comment}
\usepackage{pdflscape}
\usepackage{soul}
\usepackage{adjustbox}
\usepackage{enumitem}
\usepackage{xcolor}
\usepackage{pifont}
\usepackage{lscape}
\usepackage{setspace}
\usepackage{hyperref}
\begin{document}

\title{A Robust Simulation Framework for Verification and Validation of Autonomous Maritime Navigation in Adverse Weather and Constrained Environments}

\author{Mayur S. Patil$^{1\dagger}$,
Nataraj Sudharsan$^{1\dagger}$,
Anthony S. Saaiby$^{2\dagger}$,
JiaChang Xing$^{1}$,
Keliang Pan$^{2}$,
Veneela Ammula$^{3}$,
Jude Tomdio$^{3}$,
Jin Wang$^{3}$,
Michael Kei$^{3}$,
Heonyong Kang$^{2}$,
Sivakumar Rathinam$^{1}$,
and Prabhakar R. Pagilla$^{1}$
}

\affil{$^1$Department of Mechanical Engineering, Texas A\&M University, College Station, USA}

\affil{$^2$Department of Ocean Engineering, Texas A\&M University, College Station, USA}

\affil{$^3$American Bureau of Shipping, Spring, TX, USA}
\affil{$^\dagger$Authors contributed equally}

\email{srathinam@tamu.edu}

\begin{abstract}
Maritime Autonomous Surface Ships (MASS) have emerged as a promising solution to enhance navigational safety, operational efficiency, and long-term cost effectiveness. However, their reliable deployment requires rigorous verification and validation (V\&V) under various environmental conditions, including extreme and safety-critical scenarios. This paper presents an enhanced virtual simulation framework to support the V\&V of MASS in realistic maritime environments, with particular emphasis on the influence of weather and bathymetry on autonomous navigation performance. The framework incorporates a high-fidelity environmental modeling suite capable of simulating adverse weather conditions such as rain, fog, and wave dynamics. The key factors that affect weather, such as rain and visibility, are parameterized to affect sea-state characteristics, perception, and sensing systems, resulting in position and velocity uncertainty, reduced visibility, and degraded situational awareness. Furthermore, high-resolution bathymetric data from major U.S. ports are integrated to enable depth-aware navigation, grounding prevention capabilities, and evaluation of vessel controllability in shallow or confined waterways. The proposed framework offers extensive configurability, enabling systematic testing in a wide spectrum of maritime conditions, including scenarios that are impractical or unsafe to replicate in real-world trials, thus supporting the V\&V of MASS. 
\end{abstract}

\section{Introduction}

The maritime transportation sector is undergoing rapid technological transformation as global shipping continues to expand \cite{UNCTAD2023}, while the availability of experienced seafarers declines. In response, MASS are widely considered as a viable solution to mitigate workforce shortages and improve navigational safety, operational efficiency, and long-term cost effectiveness~\cite{wrobel2017}. To facilitate this development, the International Maritime Organization (IMO) has initiated regulatory scoping exercises to evaluate the safe integration of MASS into international shipping \cite{IMO2021}. However, despite these anticipated benefits, the reliable deployment of MASS depends on their ability to operate safely under complex and adverse environmental conditions~\cite{saaiby2026stpa}.

Maritime accident statistics highlight the severity of this challenge for introducing autonomy functions into maritime operations. Reports from the European Maritime Safety Agency (EMSA) between 2011 and 2024 indicate that more than 60\% of maritime accidents involve collisions, loss of control, groundings, or contact events. Furthermore, nearly 80\% of grounding and contact incidents, along with approximately 70\% of fatalities, are associated with severe weather conditions \cite{eliopoulou2023statistical}. General cargo ships exhibit the highest vulnerability in such scenarios, followed by Ro-Ro ferries, bulk carriers, and tankers \cite{papanikolaou2015energy}. These occurrences highlight the need for a rigorous evaluation of autonomous navigation systems in realistic and extreme operating conditions prior to large-scale deployment.

Although simulation platforms are widely used in the development of autonomous maritime systems, many existing frameworks emphasize nominal operating conditions and simplified environmental representations~\cite{hasan2023predictive,heins2017design}. Although such platforms are valuable for algorithm development and preliminary validation, a comprehensive integration of adverse weather effects and environmental constraints remains limited. Furthermore, investigations on the impact of autonomous ship technologies highlight that verification and validation under realistic environmental uncertainty continues to be a critical open challenge~\cite{chan2025investigating}. These limitations restrict the ability to examine coupled interactions between environmental disturbances, sensor degradation, vessel maneuverability, and autonomous navigation algorithms, particularly in ports and coastal regions where operational margins are reduced.

To address this gap, this work presents an enhanced virtual simulation framework that builds on the existing platform jointly developed by Texas A\&M University and American Bureau of Shipping (ABS) in~\cite{patil2025virtual}. The expanded framework integrates an environmental modeling suite that captures a spectrum of adverse weather conditions, including fog, rain, wind, and wave dynamics, rendered with high physical fidelity. The severity of the weather is parameterized to provide extensive user configurability, allowing controlled adjustment of environmental conditions to evaluate the robustness of MASS in a wide range of operational scenarios. These weather variations directly influence sea-state characteristics and sensor performance, enabling systematic assessment of vessel performance under heightened operational risk and impaired perception and situational awareness. Additionally, the framework incorporates high-resolution bathymetric data from major U.S. ports, including the Port of Houston and the Port of Los Angeles, to support geographically realistic testing scenarios. Bathymetric data are integrated to evaluate depth-aware navigation and grounding avoidance~\cite{patil2026colregs}, complemented by high-resolution visualization within the simulation environment to facilitate the user during testing. Real-world underwater terrain profiles allow evaluation of vessel controllability and safety margins in shallow waterways, strengthening the realism of the test cases.

Simulations were conducted using baseline perception, planning, and control algorithms to examine the impact of environmental conditions on vessel behavior. A set of integrated performance indicators (PI) were employed to quantitatively evaluate collision avoidance behavior and controller robustness under uncertain environmental conditions. The results confirm that environmental factors substantially affect vessel performance, highlighting the critical importance of incorporating realistic environmental variability into V\&V processes for MASS. 

\section{Preliminaries}
This section provides a review of the simulation framework and V\&V procedure presented in~\cite{patil2025virtual}, which forms a basis for the enhancements and extensions introduced in the present work.

\subsection {Simulation Framework}
\label{section 2.1}
\begin{figure}[h]
    \centering
    \includegraphics[width=0.6\textwidth]{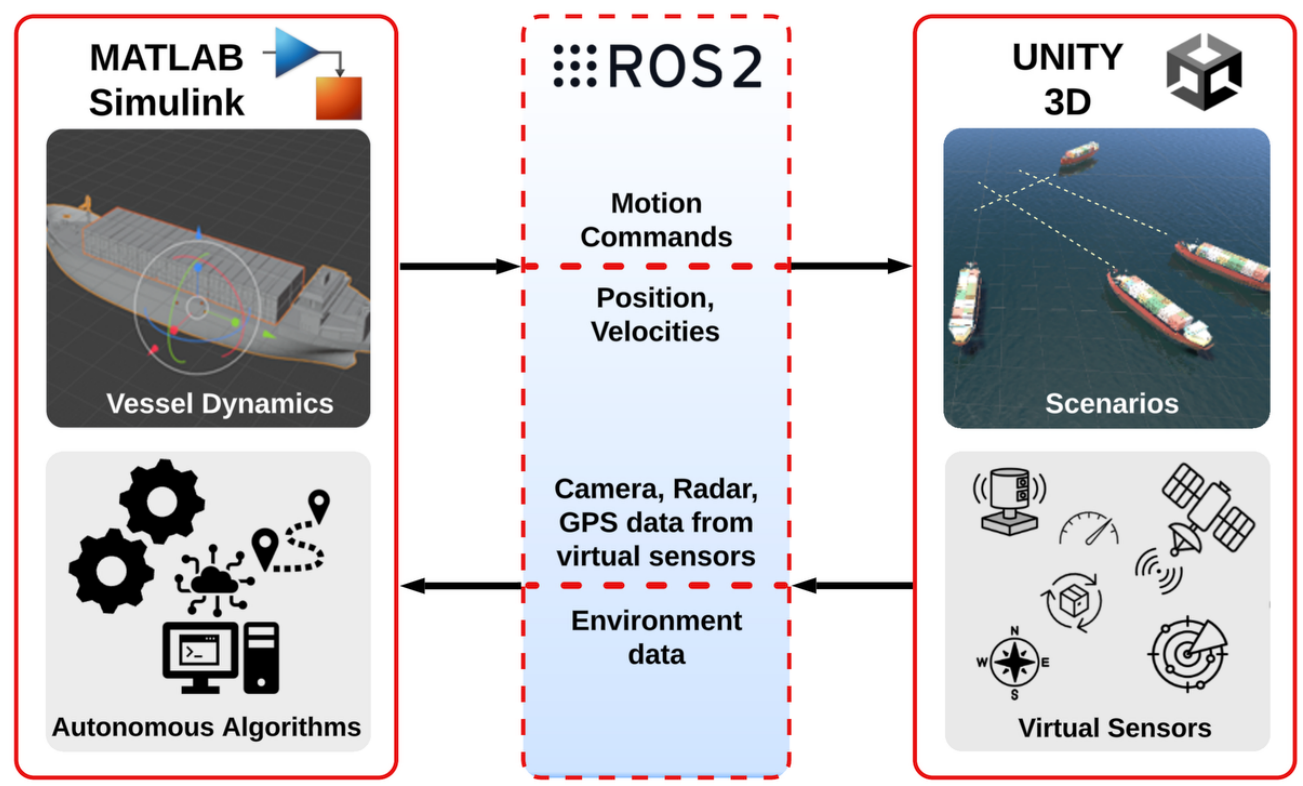}
    \caption{Simulation Framework Architecture (adopted from \cite{patil2025virtual})}
    \label{Simulation_Framework_Architecture}
\end{figure}

The simulation framework is an integrated platform that creates digital twins of the real-world maritime environment in which autonomous navigation strategies are tested in various maritime navigation scenarios. It is built on three platforms: Unity, MATLAB-Simulink, and ROS2. The architecture, shown in Figure \ref{Simulation_Framework_Architecture}, illustrates how these platforms have been structured to synchronously work together. In this architecture, Unity provides the visualization of the maritime environment and emulates a set of onboard maritime sensors, such as GPS and LiDAR; MATLAB Simulink, functioning as a computation center, processes sensor data received from Unity and outputs vessel motion back to Unity for visualization; and ROS2 is the interface that enables communication between Unity and MATLAB Simulink.

In this architecture, the autonomous navigation algorithm is developed using eight modules (see Fig. 2 in~\cite{patil2025virtual}). The Scenario module is implemented within Unity, which generates navigation scenarios, and the remaining modules are implemented in MATLAB. Following this modular structure, the Unity Scenario module first generates a navigation scenario where onboard sensors are also modeled. The Sensor Fusion module then takes sensor data as input, providing an estimation of the position and velocity of the vessel of interest and its surrounding objects. The Situational Awareness module uses this information to assess if evasive maneuvering is needed based on a set of pre-defined collision thresholds, such as the time to closest point of approach (TCPA) and  distance at closest point of approach (DCPA). The Planning module, consisting of two components: a global planner and a Collision Detection and Collision Avoidance (CDCA) planner, determines the desired trajectory of the vessel of interest. The global planner computes the feasible shortest trajectory toward the destination without accounting for other vessels, while the CDCA planner initiates a set of rerouting waypoints that avoid potential collisions when triggered by the Situational Awareness module. If evasive maneuvering is required, the Guidance module generates a continuous trajectory using this set of rerouting waypoints. The Control module then sends control signals to the rudder and propeller, regulating the ship's heading and speed to follow the rerouted trajectory. Finally, the Vessel Dynamics module models the motion of the vessel, replicating vessel responses to control commands and external environmental loads. This simulation framework features real-time synchronization between the vessel's response to the simulated maritime environment, which is realized by continuous interaction between Simulink and Unity.

\subsection{Verification and Validation}    

The work presented in \cite{patil2025virtual} establishes a methodology for V\&V of autonomous navigation algorithms through a set of benchmark test scenarios and PIs. The 22 IMAZU scenarios were selected due to their widespread adoption in autonomous collision avoidance research \cite{cai2013evaluating}. These scenarios capture high-collision-risk situations commonly encountered in routine maritime operations, providing a rigorous basis for evaluating autonomous collision avoidance strategies. The associated PIs include quantitative measures such as minimum passing distance (MPD) and root-mean-square errors in speed ($RMSE_{V}$) and heading angle ($RMSE_{\psi}$) as described in~\cite{patil2025virtual}.


\section{Environmental Modeling}
Reliable environmental modeling is essential for the V\&V of autonomous maritime systems. Environmental conditions influence vessel motion and sensing reliability and must be realistically represented within a robust virtual testing framework. Consequently, environmental effects in this work are modeled through two key components: weather, which impacts sensor performance, and bathymetry, which defines navigational constraints and safe operating regions.

\subsection{Weather Modeling}
Weather affects autonomous maritime navigation primarily by degrading onboard sensing, particularly marine radar, which provides long range perception for collision avoidance and situational awareness. Since most navigation algorithms rely heavily on radar measurements to track surrounding traffic, weather-induced performance degradation can directly reduce perception reliability, destabilize tracking, and affect downstream decision making. A robust virtual testing framework must therefore capture weather conditions and their functional impact on radar sensing. Although explicit electromagnetic (EM) wave propagation models can represent these effects with high fidelity, they introduce significant computational overhead with limited benefit for evaluating high-level autonomy behavior. For large scale scenario testing and real-time simulation, a computationally efficient representation that preserves the dominant operational effects of weather on radar performance is more appropriate. Consequently, the framework integrates a configurable radar sensing model for real-time execution with a physics-informed weather model that captures environmental variability and its impact on radar performance.

\subsubsection{Radar Sensor Modeling:}

To enable real-time and large scale simulation, the radar sensor is modeled using a computationally efficient method that produces discrete target detections based on geometric visibility and configurable sensing parameters. At each update cycle, potential targets within the sensing volume are identified using Unity’s \textit{OverlapSphere} query centered at the radar location, with a radius equal to the maximum detection range. This proximity based filtering limits processing to relevant objects. For each candidate, a small set of raycasts is performed to verify the direct line-of-sight between the radar and the target. Objects occluded by other vessels, terrain, or infrastructure are rejected, allowing the model to represent occlusion effects that are critical in maritime environments. For each visible object, the sensor produces measurements of relative range and bearing in the ego vessel frame. The target velocity is estimated using a sliding window over recent relative position measurements, providing stable motion estimates while maintaining low computational overhead. A key feature of the model is its high degree of configurability. All sensor characteristics are parameterized through a graphical interface in Unity, including operating frequency, maximum detection range, update rate, scan frequency, and tracking window duration. This enables the representation of different marine radar classes, such as X-band systems with higher spatial resolution and S-band systems with longer range and improved performance in adverse weather. The configurable design allows for the evaluation of the autonomy performance in various sensor configurations and operating modes.

\subsubsection{Weather Effects On Radar:}
\label{sec:weather_condition_modeling}

The proposed weather module is designed to achieve two complementary objectives: (i) controlled generation of diverse environmental conditions, and (ii) physics informed modeling of weather induced sensor degradation.

First, weather conditions are represented using three independent severity variables: rain, fog, and sea state. Each variable is defined on a normalized scale from 0 to 10 and assigned to physically meaningful quantities. The severity of rain spans from $0~\mathrm{mm/h}$ to $100~\mathrm{mm/h}$, the severity of fog is mapped to the liquid water content ($\mathrm{g/m^3}$) and the corresponding visibility levels, and severity of the sea state is associated with significant wave height and peak period. These mappings are configurable, enabling adaptation to different operational environments and experimental requirements. The continuous representation supports systematic testing across nominal to extreme conditions. Furthermore, a graphical user interface is designed within Unity that enables real-time adjustment of weather severity values. Rain controls particle emission rate, fog modifies scene visibility, and the sea state adjusts the roughness of the ocean surface. This interactive capability enables rapid scenario configuration with immediate visual feedback.

Second, the functional impact of weather on radar performance is modeled through frequency dependent atmospheric attenuation models. Specifically, the attenuation due to rain is calculated using the ITU-R P.838 model \cite{ITU838}, where the specific attenuation is given by
\begin{equation}
\gamma_r = k(f) R^{\alpha(f)} \quad \text{(dB/km)},
\end{equation}
with $R$ representing the rain rate (mm/hr) and coefficients $k$ and $\alpha$ determined by the operating frequency. This formulation captures the nonlinear increase in attenuation with rain intensity and its stronger impact at higher frequencies. Similarly, Fog attenuation is modeled using the ITU-R P.840 recommendation \cite{ITU840} as
\begin{equation}
\gamma_f = K_{\ell}(f,T) M \quad \text{(dB/km)},
\end{equation}
where $M$ denotes the volume of liquid water content in the air. Fog-induced losses are mainly due to absorption and become more significant at higher frequencies. The total atmospheric attenuation ($A_{w}$) is computed as the combined effect of rain and fog and applied as a distance-dependent propagation loss for a path length $d$ as
\begin{equation}
A_{w} = (\gamma_r + \gamma_f) d \quad \text{(dB)}.
\end{equation}

$A_{w}$ reduces the power of the returned echo and therefore lowers the signal-to-noise ratio (SNR) at the receiver. The SNR represents the ratio of the received target echo power to the total noise power at the receiver, which is evaluated using the monostatic radar equation given in \cite{Skolnik}:
\begin{equation}
\mathrm{SNR} =
\frac{P_t G^2 \lambda^2 \sigma}
{(4\pi)^3 R^4 \, k T_0 B F \, L_{\text{w}}},
\end{equation}
where $P_t$ is the transmit power, $G$ is the antenna gain, $\lambda$ is the wavelength, $\sigma$ is the radar cross section (RCS) of the target, and $R$ is the target range. The term $kT_0B$ represents the thermal noise power, $F$ is the receiver noise profile, and $L_{\text{w}} = 10^{A_{\text{w}}/10}$ denotes the linear propagation loss due to weather. Consequently, increasing environmental severity increases the overall propagation loss, thus reducing the received echo power and the SNR. The SNR affects detection capability and the accuracy of the estimated target. At low SNR, noise has a stronger influence on the received echo, distorting the wave and increasing the uncertainty in the estimated arrival time. As the SNR increases, the effect of noise is reduced, and the range estimation accuracy improves.
As shown in \cite{Curry}, the standard deviation of the range estimate is given by
\begin{equation}
\sigma_R = \frac{c}{2B\sqrt{2\mathrm{SNR}}},
\end{equation}
where $c$ is the speed of light and $B$ is the receiver bandwidth.  Furthermore, the accuracy of the velocity estimate obtained from Doppler effect is similarly expressed as
\begin{equation}
\sigma_v = \frac{\lambda}{2 T_{\mathrm{dwell}} \sqrt{2\,\mathrm{SNR}}},
\end{equation}
where $T_{\mathrm{dwell}}$ is dwell time. A longer dwell time improves frequency estimation by increasing the effective observation interval, while shorter wavelengths increase sensitivity to target motion. The computed standard deviations are used to parameterize zero mean Gaussian distribution applied to the true target position and velocity. As a result, increased weather severity produces higher atmospheric attenuation, reduced SNR, and a corresponding increase in measurement variance; establishing a physically consistent relationship between environmental conditions and their influence on radar accuracy.

\subsection{Bathymetry Modeling}

A physically consistent representation of the maritime environment requires consideration of subsurface constraints. In this work, bathymetric modeling is incorporated as a core element of the virtual testing framework to capture spatially varying water depth and seabed topology. Most existing autonomous navigation approaches emphasize collision avoidance at the surface-level and treat the operational domain as a two-dimensional free space. This abstraction ignores depth variability and under-keel clearance constraints, despite their critical influence on safety, maneuverability, and hydrodynamic loading. For MASS operating in infrastructure-dense or shallow-water environments, bathymetric effects must therefore be explicitly represented within the autonomy validation framework. This integration extends the capabilities of the virtual testing framework by supporting: (i) high-fidelity three-dimensional seabed reconstruction within the Unity-based simulation environment; (ii) depth-aware navigation through the explicit representation of non-navigable shallow regions as static environmental constraints; and (iii) incorporation of depth-dependent effects within hydrodynamic wave-load modeling to capture shallow-water influences on vessel response.

\subsubsection{High-Fidelity Unity Visualization:}

The generation of bathymetry-based terrain within the Unity simulation environment is implemented through a structured, multi-stage processing pipeline designed to preserve geospatial fidelity while ensuring compatibility with Unity rendering constraints. The workflow consists of several steps: hydrographic data acquisition, spatial tiling, preprocessing, heightmap conversion, and terrain integration.

Bathymetric datasets are acquired from authoritative hydrographic agencies, primarily the National Oceanic and Atmospheric Administration (NOAA) and the U.S. Geological Survey (USGS). However, Public data distribution policies restrict raster downloads to approximately $10{,}000 \times 10{,}000$ pixels per file. For large geographic regions, this limitation necessitates either down-sampling or reduced-resolution data, which can introduce discretization errors when directly imported into Unity. As a result, a hierarchical spatial subdivision strategy is employed to generate bathymetry based terrains. The operational area is subdivided into uniformly sized bathymetric tiles prior to acquisition. Each tile is processed independently to comply with raster size limitations while preserving a higher effective spatial resolution. During Unity integration, tiles are mapped to corresponding terrain segments using consistent spatial scaling parameters, enabling seamless stitching across adjacent terrain sections.

\begin{figure}[tbh!]
    \centering
    \includegraphics[width=0.8\linewidth] {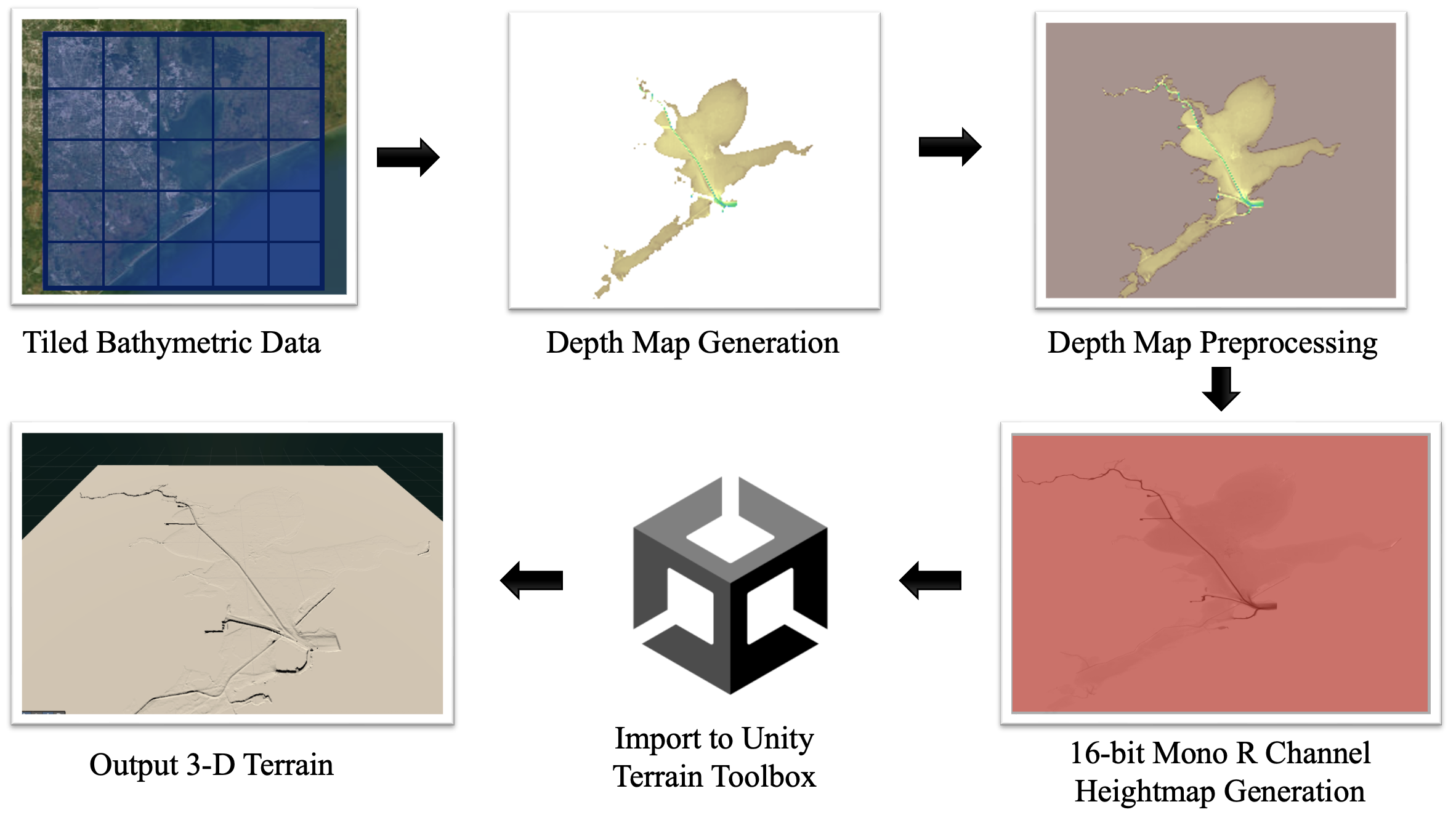}
    \caption{Bathymetry-based terrain generation in Unity}
    \label{fig:unity_bathymetry_pipeline_red}
\end{figure}

Bathymetric datasets are typically provided in GeoTIFF (.tiff) format containing georeferenced elevation values. A set of preprocessing steps is  performed on these datasets to extract depth information, merge the datasets, remove regions outside the defined area of interest, and mitigate discontinuities arising from interpolation or survey overlap. The resulting elevation model is then converted into 16-bit mono-channel (R-channel) height maps compatible with Unity’s Terrain Toolbox. Compared to conventional 8-bit grayscale encoding, 16-bit depth representation significantly increases vertical resolution, reducing quantization errors and minimizing artificial terracing effects in shallow and transitional regions. These height maps are then resized and imported into Unity with the  appropriate resolution parameters using Unity Terrain Toolbox to accurately represent seabed contours, dredged navigation channels, and coastal morphology suitable for high-fidelity maritime simulation. Figure~\ref{fig:unity_bathymetry_pipeline_red} graphically illustrates the complete bathymetric terrain generation workflow. 

This methodology was applied to two major U.S. ports: the Port of Houston and the Port of Los Angeles. These environments were selected due to their complex navigational geometries, shallow-water constraints, dredged channels, and high vessel traffic density, characteristics that are critical to the evaluation of autonomous maritime navigation systems.

\subsubsection{Depth-Aware Planning and Decision Making:}

Bathymetric data are directly integrated into the autonomous navigation stack to enable depth-aware planning. Unlike conventional approaches that treat bathymetry as contextual information, the proposed framework formalizes the seabed topology as a structural constraint that governs the motion feasibility of the vessel.

First, the bathymetric GeoTIFF datasets are transformed into shallow-water occupancy grids to enable explicit depth-constrained planning. Georeferenced depth matrices are extracted and converted into MATLAB-compatible formats while preserving the spatial tiling structure established during terrain reconstruction. A vessel-dependent minimum safe depth threshold derived from draft and required under-keel clearance is applied to classify grid cells as navigable or non-navigable. The resulting binary maps are spatially concatenated to form a unified large-scale occupancy grid that represents constraints based on depth in the operational domain. Second, this occupancy representation is integrated into the velocity obstacle based collision detection and collision avoidance (CDCA) framework. Shallow-water regions are modeled as static obstacles, enabling grounding avoidance consideration within the same decision-making structure that handles dynamic vessel collision avoidance.



\subsubsection{Wave Load Calculations:}

Bathymetric data are  also incorporated into the wave load computation to account for depth-dependent hydrodynamic effects. During simulation, the bathymetric dataset is continuously queried to provide the local water depth estimates in real-time to the wave models implemented within the MATLAB/Simulink environment. This retrieved depth information is utilized in the wave spectrum generation, enabling an accurate representation of wave characteristics under varying seabed conditions. This facilitates the modeling of depth-influenced wave transformation processes, which captures induced wave loads acting on the vessel due to the varied sea topography within the simulation framework.

\begin{figure}[h]
    \centering
    \includegraphics[width=0.75\textwidth]{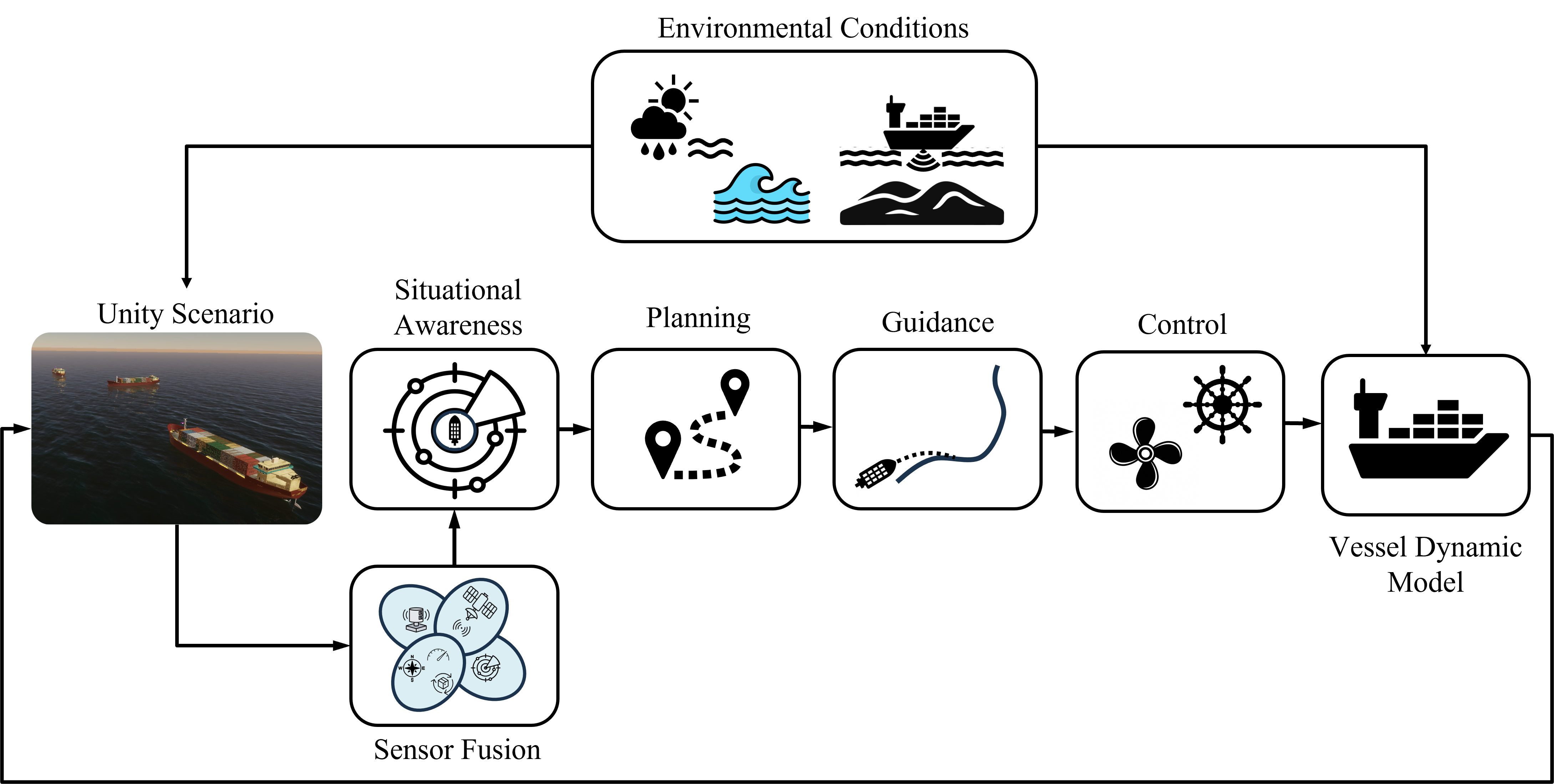}
    \caption{Enhanced modular simulation framework}
    \label{Simulation_Framework_Modulation}
\end{figure}

\subsection{Enhanced Simulation Framework}

\begin{figure}[tbh!]
    \centering
    \begin{subfigure}[b]{0.47\textwidth}
        \centering
        \includegraphics[width=\textwidth]{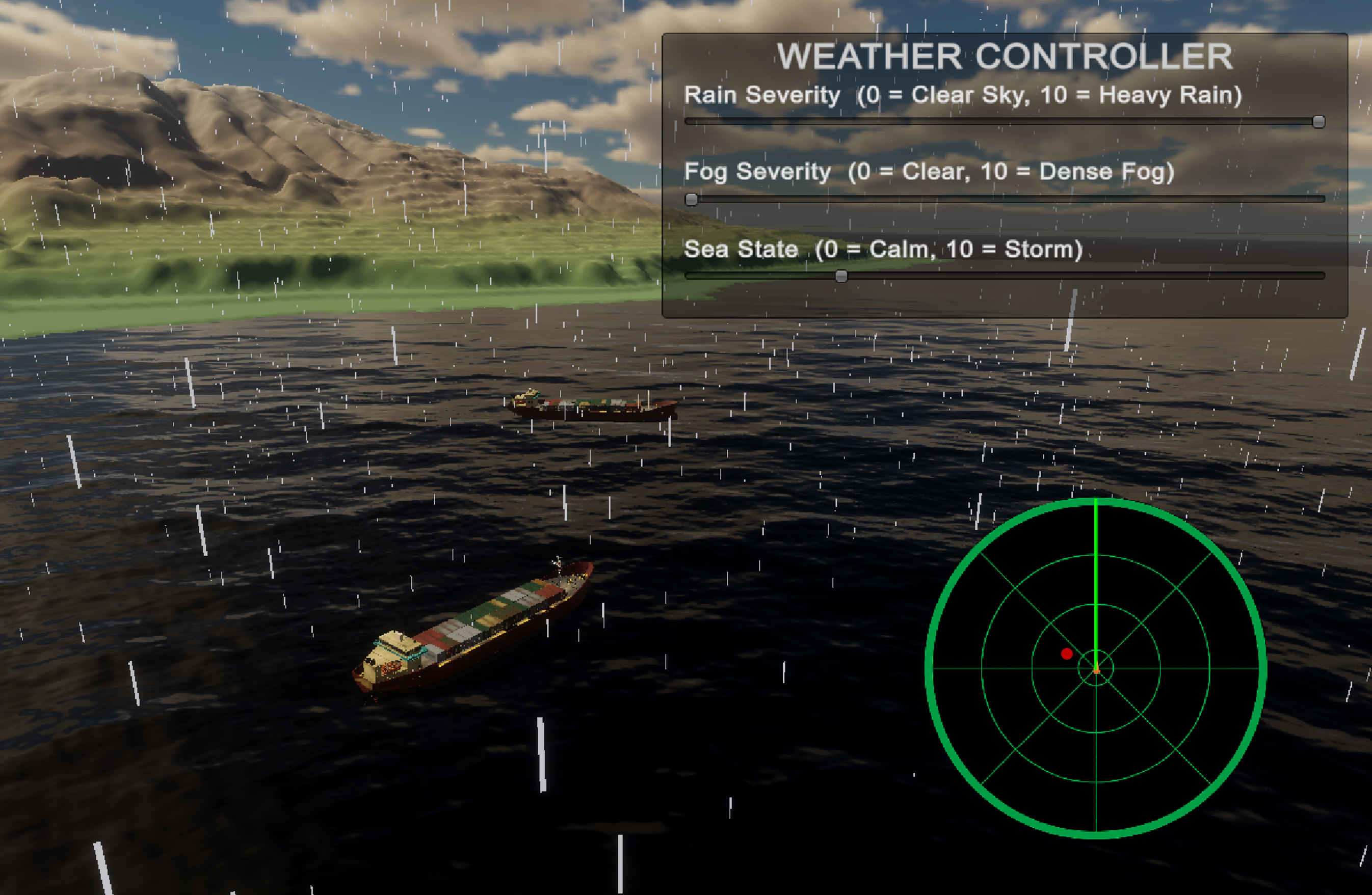}
        \caption{}
        \label{fig:weather}
    \end{subfigure}
    \hfill 
    \begin{subfigure}[b]{0.47\textwidth}
        \centering
        \includegraphics[width=\textwidth]{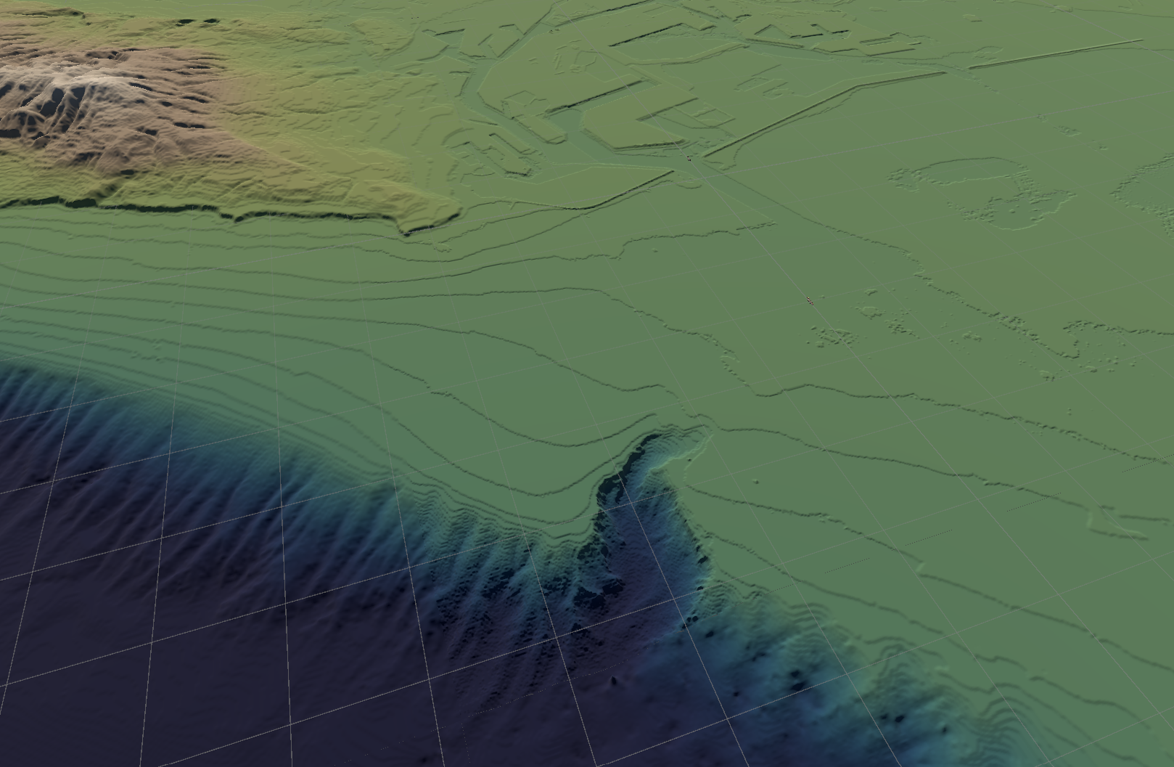}
        \caption{}
        \label{fig:bathy}
    \end{subfigure}
    \caption{ Unity visualizations (a)Weather severity (b) Bathymetry representation. }
    \label{fig:unity_visuals}
\end{figure}

Figure~\ref{Simulation_Framework_Modulation} illustrates the enhanced simulation framework, which incorporates a realistic bathymetric profile of the operating environment together with key environmental conditions such as fog-induced visibility reduction, rainfall and  wave conditions. Figure~\ref{fig:unity_visuals} presents the corresponding visualization in Unity environment, where the scene includes  rainfall and detailed seabed topography, providing a visual representation of the environmental complexity introduced in the simulation model. This visualization layer supports qualitative evaluation and validation of the simulated conditions within an interactive 3D environment.


\section{Results and Discussion }

To evaluate the influence of environmental conditions on the autonomous navigation system's performance, simulations were conducted within a realistic maritime environment. The modeled environment represents the Port of Los Angeles, defined by latitude bounds [33.5666°, 33.7780°] and longitude bounds [-118.2838°, -118.0318°], covering approximately $5.48 \times 10^{8}$ m$^{2}$ (23.3 km × 23.5 km). Vessel traffic consisted of an Imazu encounter scenario with two target vessels: vessel V1 initialized at (3500 m, 3500 m) with a heading of 270°, representing a crossing vessel, and vessel V2 initialized at (0 m, 7000 m) with a heading of 180°, representing a head-on vessel. All vessels were initialized with a nominal speed of 10 m/s. The speeds and headings of the non-autonomous (target) vessels were predefined, while only the ego (autonomous) vessel was subject to closed-loop control of speed and heading. All vessels were modeled as S175 containerships with dimensions of length ($L$) 175.0 m, beam ($B$) 25.4 m, and ($D$) draft 9.5 m. In the first study, a fixed nominal radar configuration (Power: 10 kW, Antenna gain: 26 dB, Bandwidth: 20 MHz, Dwell time: 8 ms) was used while varying environmental conditions. Three representative weather scenarios were considered: (i) ideal conditions (0 mm/hr rain, 0 g/m$^{3}$ atmospheric water density, 0.5 m significant wave height), (ii) moderate conditions (50 mm/hr rain, 0.25 g/m$^{3}$ water density, 1 m wave height), and (iii) severe conditions (100 mm/hr rain, 0.5 g/m$^{3}$ water density, 1.5 m wave height), representing low, intermediate, and extreme disturbances. In the second study, severe weather conditions were fixed while varying radar sensor configurations to capture the effect of sensing capabilities on autonomous navigation. The radar configurations considered were: (i) a high-power radar (25 kW, 30 dB gain, 28 MHz bandwidth, 10 ms dwell time), (ii) a nominal radar (10 kW, 26 dB, 20 MHz, 8 ms), and (iii) a low-power radar (7 kW, 22 dB, 10 MHz, 5 ms). This evaluation enables systematic isolation of environmental effects on the autonomy performance.

\begin{figure}[t]
\centering
\begin{subfigure}[t]{0.48\linewidth}
    \centering
    \includegraphics[width=\linewidth]{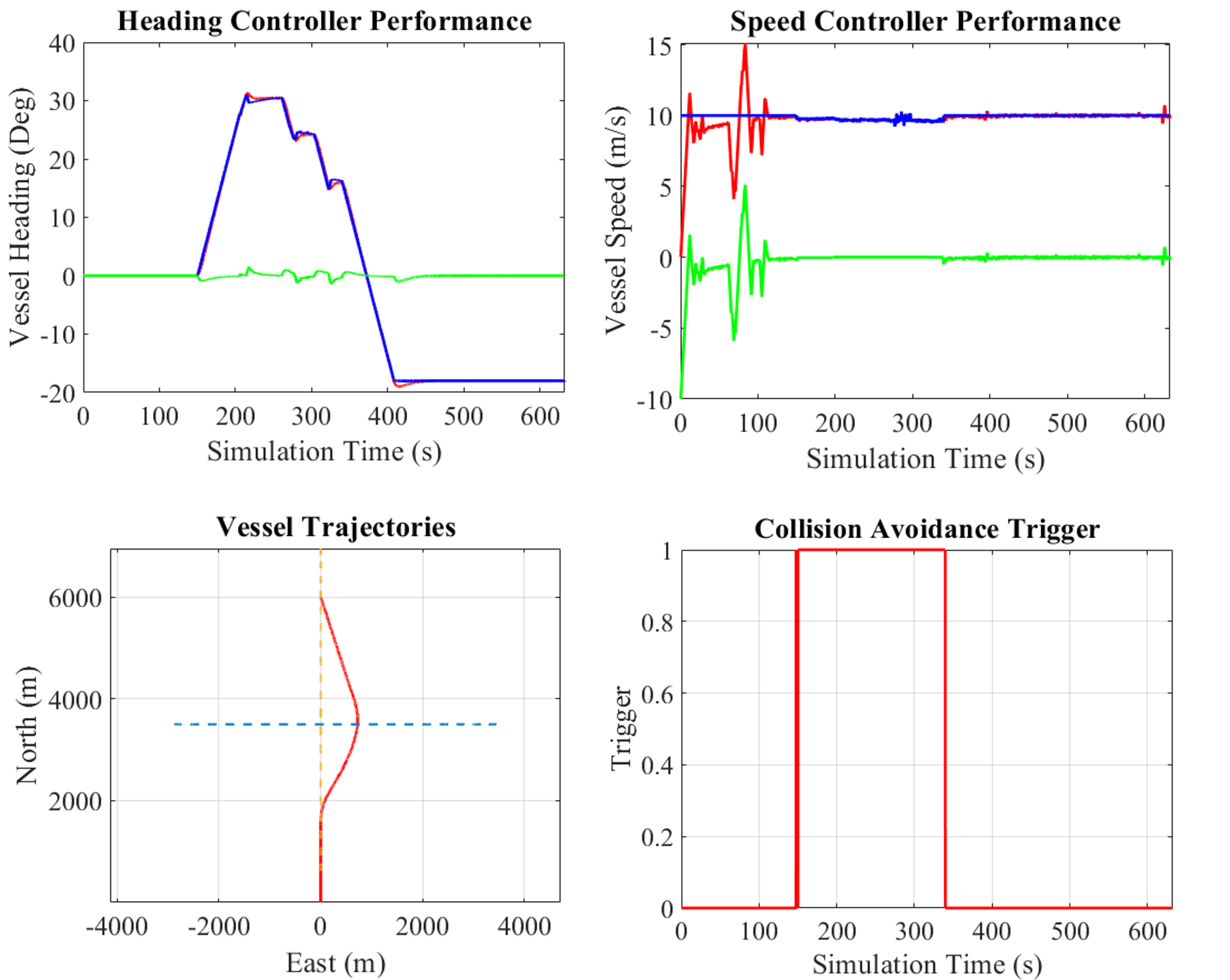}
    \caption{Ideal weather}
    \label{fig:good_weather}
\end{subfigure}
\hfill
\begin{subfigure}[t]{0.48\linewidth}
    \centering
    \includegraphics[width=\linewidth]{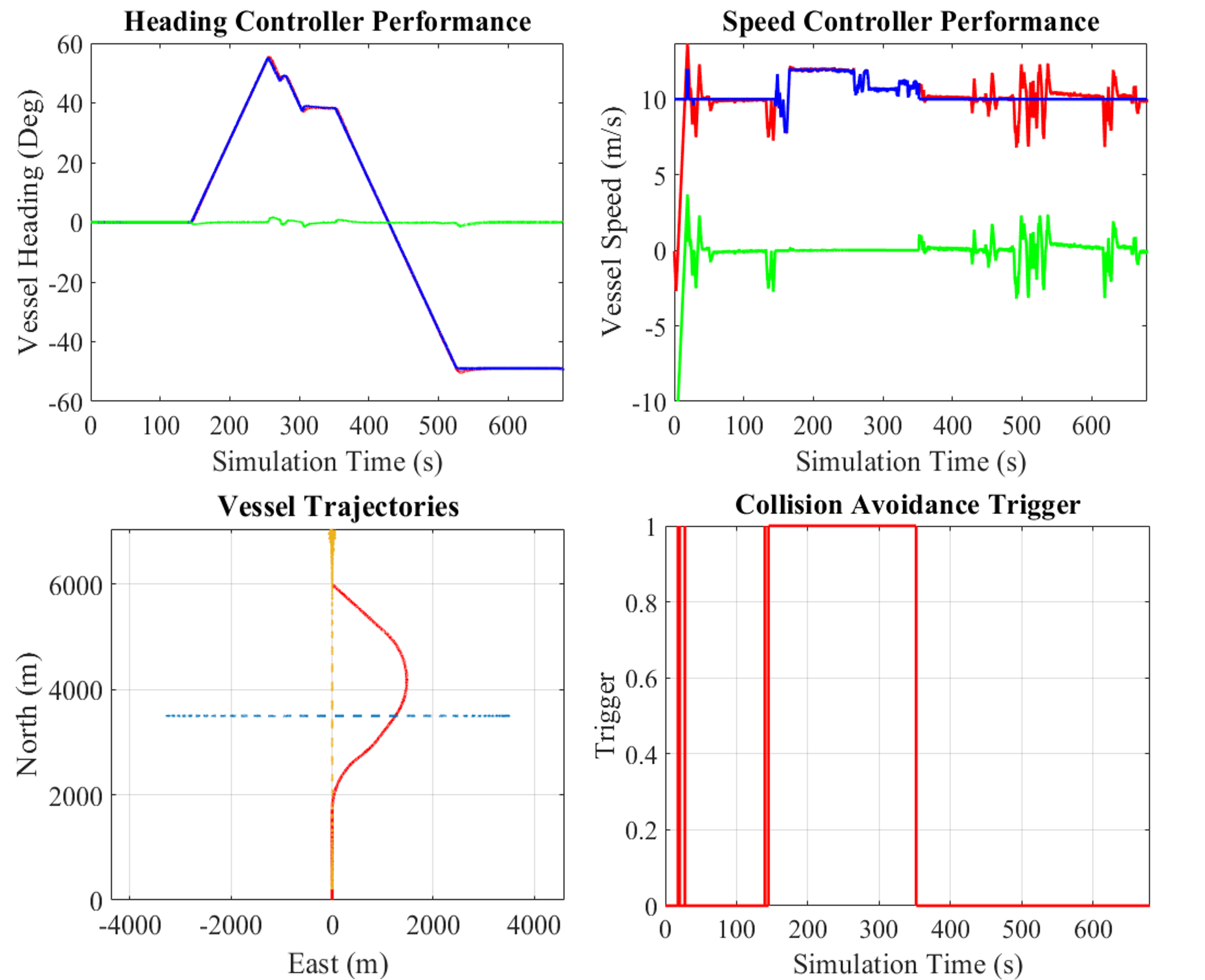}
    \caption{Severe weather}
    \label{fig:bad_weather}
\end{subfigure}

\caption{Controller performance, vessel trajectories, and collision avoidance trigger from the situational awareness module under (a) ideal and (b) severe weather using nominal sensor.}
\label{fig:weather_comparison}
\end{figure}

Figure~\ref{fig:weather_comparison} compares controller performance and vessel trajectories under a nominal radar configuration for ideal and severe weather conditions. Under ideal weather, the heading and speed controllers achieve tracking errors of 0.343$^\circ$ and 1.1 m/s (2.12 knots), respectively, increasing to 0.355$^\circ$ and 1.4 m/s (2.72 knots) in severe weather. Trajectory analysis reveals that radar-induced uncertainty in target vessel position and velocity significantly increases in adverse conditions. Specifically, positional and velocity uncertainties are approximately 0.1 m and 0.02 m/s in ideal weather, whereas in severe weather they increase to approximately $\pm$60 m and $\pm$15 m/s, respectively, gradually reducing as inter-vessel distance decreases. Correspondingly, the ego vessel exhibits a peak heading deviation of 31.5$^\circ$ in ideal weather compared to 55$^\circ$ in severe weather, resulting in a longer traversed path in the latter case. This deviation is attributed to the behavior of the situational awareness module, which generates collision avoidance triggers based on instantaneous DCPA and TCPA estimates derived from target vessel state measurements. In severe weather, increased sensing uncertainty leads to oscillatory triggering, as evidenced by intermittent activations at 20.15 s, 21.55 s, 22.35 s, and 29.75 s as seen in collision avoidance trigger plot, despite vessels being outside the nominal collision zone. These premature and fluctuating triggers cause earlier avoidance maneuvers, leading to suboptimal trajectory decisions. Although greater separation is ultimately maintained, this outcome results from uncertainty-induced behavior rather than deliberate planning and may become hazardous in high-traffic environments where unnecessary maneuvers could escalate encounter complexity. Table~\ref{tab:weather_performance} summarizes the PIs for the first study. The trend observed validates degradation in vessel performance under severe weather conditions, primarily attributable to the premature activation of collision avoidance triggers in response to increased uncertainty in target vessel state estimates.

\begin{table}[h]
\centering
\small
\begin{tabular}{|l|c|c|c|c|}
\hline
\textbf{Weather} & 
\textbf{MPD V1 [m]} & 
\textbf{MPD V2 [m]} & 
\textbf{$RMSE_{V}$ [m/s]} & 
\textbf{$RMSE_{\psi}$ [deg]} \\
\hline
Ideal     & 606  & 701   & 1.1 & 0.343 \\
\hline
Moderate  & 607 & 784 & 1.1 & 0.350 \\
\hline
Severe    & 610  & 1137 & 1.4  & 0.355  \\
\hline
\end{tabular}
\caption{Performance Indicators under different weather conditions.}
\label{tab:weather_performance}
\end{table}

\begin{figure}[t]
\centering
\begin{subfigure}[t]{0.47 \textwidth}
    \centering
    \includegraphics[width=\linewidth]{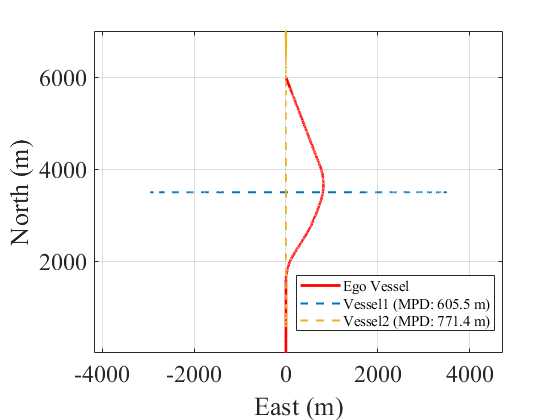}
    \caption{High-power radar}
    \label{fig:good_sensor_bad_weather}
\end{subfigure}
\hfill
\begin{subfigure}[t]{0.47\textwidth}
    \centering
    \includegraphics[width=\linewidth]{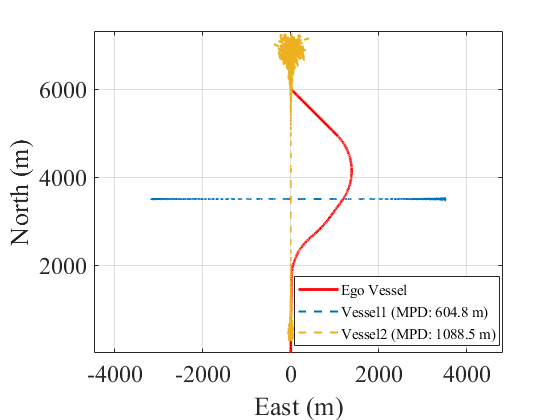}
    \caption{Low-power radar}
    \label{fig:bad_sensor_bad_weather}
\end{subfigure}

\caption{Controller performance and trajectory under severe weather conditions for (a) high-power radar and (b) low-power radar configurations.}
\label{fig:sensor_comparison}
\end{figure}

Figure~\ref{fig:sensor_comparison} presents a comparison of vessel trajectories under severe weather conditions for high-power and low-power radar configurations. Under these conditions, the ego vessel equipped with the high-power radar exhibits reduced uncertainty in the estimation of target vessel positions and velocities, resulting in stable performance across autonomy modules.
In contrast, the low-power radar configuration leads to increased trajectory irregularities and visibly noisier position and velocity estimates of target vessels. The orange cluster observed near the head-on vessel in Figure~\ref{fig:bad_sensor_bad_weather} represents noisy radar measurement points under severe weather, illustrating degraded resolution and elevated measurement uncertainty. This increased uncertainty propagates through the situational awareness and planning modules, ultimately affecting control responses. Specifically, erroneous measurements induce significantly higher oscillatory collision triggers compared to the first study, resulting in noisier desired trajectories generated by the decision-making algorithm and consequently degrading overall vessel performance. Table~\ref{tab:radar_performance} summarizes the PIs for the second study, reflecting the effect of reduced radar power on tracking accuracy, compromising decision making under adverse weather.
Nevertheless, safe separation is consistently maintained, and effective collision avoidance is achieved, demonstrating the robustness of the baseline autonomy algorithms used in the autonomy stack of the ego vessel in compensating for sensing degradation while preserving safe and efficient operation. Altogether, the simulation framework provides a systematic means to assess the capabilities and limitations of autonomous modules, hence facilitates V\&V autonomous navigation functionalities.

\begin{table}[h]
\centering
\small
\begin{tabular}{|l|c|c|c|c|}
\hline
\textbf{Radar} & 
\textbf{MPD V1 [m]} & 
\textbf{MPD V2 [m]} & 
\textbf{$RMSE_{V}$ [m/s]} & 
\textbf{$RMSE_{\psi}$ [deg]} \\
\hline
High-power     & 606  & 771  & 1.3 & 0.347 \\
\hline
Nominal Power  & 610  & 1137 & 1.4 & 0.355 \\
\hline
Low-power  & 605  & 1089 & 1.5 & 0.412 \\
\hline
\end{tabular}
\caption{Performance indicators under varying radar sensor configurations.}
\label{tab:radar_performance}
\end{table}
\section{Conclusion}

This paper presented an enhanced virtual simulation framework to support rigorous V\&V of MASS in realistic and safety-critical maritime environments. The framework extends prior efforts by incorporating configurable, high-fidelity weather models and representative bathymetric data from major U.S. ports, enabling comprehensive environment-aware testing. The framework establishes a physics-consistent relationship between environmental conditions and radar sensing performance by explicitly modeling propagation loss and signal degradation under adverse weather. In addition, bathymetric integration enables the representation of wave-induced loads and vessel responses influenced by varying seabed topography, providing a more complete characterization of vessel–environment interactions.

A set of simulation tests were conducted across diverse weather conditions and radar configurations to evaluate vessel performance using integrated PIs. The findings highlight a clear sensitivity of autonomous vessel performance to adverse weather conditions, and varied seabed topography, reinforcing the necessity of environment-aware testing in MASS V\&V processes. Overall, the framework enables repeatable and scalable testing in complex maritime conditions to supports extended V\&V of MASS, enhancing their credibility before reliable and safe large-scale deployment.

\section*{Acknowledgment}
This project is supported by the American Bureau of Shipping (ABS).


\bibliographystyle{IEEEtran}
\bibliography{references}

@report{UNCTAD2023,
  title={Review of Maritime Transport 2023},
  author={UNCTAD},
  year={2023},
  institution={United Nations Conference on Trade and Development}
}

@misc{IMO2021,
  title={Regulatory Scoping Exercise for the Use of Maritime Autonomous Surface Ships (MASS)},
  author={{International Maritime Organization}},
  year={2021},
  howpublished={IMO MSC.1/Circ.1638}
}

@inproceedings{patil2025virtual,
  title={Virtual Framework for Verification and Validation of Marine Autonomous Navigation},
  author={Patil, Mayur S and Sudharsan, Nataraj and Saaiby, Anthony and Xing, JiaChang and Chan, Jevon and Ammula, Veneela and Tomdio, Jude and Wang, Jin and Rathinam, Sivakumar and Pagilla, Prabhakar and others},
  booktitle={SNAME Maritime Convention},
  year={2025},
  organization={SNAME}
}

@unpublished{patil2026colregs,
  author = {Patil, Mayur S and Sudharsan, Nataraj and others},
  title  = {COLREGs Compliant Collision Avoidance and Grounding Prevention for Autonomous Marine Navigation},
  year   = {2026},
  note   = {Manuscript submitted for publication}
}

@unpublished{saaiby2026stpa,
  author = {Saaiby, Anthony and Patil, Mayur S and others},
  title  = {A framework for identification and prioritization of critical
factors in transition between manual and autonomous navigation
functions in marine vessels},
  year   = {2026},
  note   = {Manuscript submitted for publication}
}

@article{eliopoulou2023statistical,
  title={Statistical analysis of accidents and review of safety level of passenger ships},
  author={Eliopoulou, Eleftheria and Alissafaki, Aimilia and Papanikolaou, Apostolos},
  journal={Journal of Marine Science and Engineering},
  volume={11},
  number={2},
  pages={410},
  year={2023},
  publisher={MDPI}
}

@inproceedings{papanikolaou2015energy,
  title={Energy efficient safe ship operation (SHOPERA)},
  author={Papanikolaou, Apostolos and Zaraphonitis, George and Bitner-Gregersen, Elzbieta and Shigunov, Vladimir and El Moctar, Ould and Soares, Carlos Guedes and Reddy, Devalapalli N and Sprenger, Florian},
  booktitle={SNAME Maritime Convention},
  pages={D021S007R013},
  year={2015},
  organization={SNAME}
}

@article{wrobel2017,
  author = {Wróbel, K. and Montewka, J. and Kujala, P.},
  title = {Towards the development of a system-theoretic model for safety assessment of autonomous ships},
  journal = {Reliability Engineering \& System Safety},
  volume = {165},
  pages = {155--169},
  year = {2017}
}

@article{heins2017design,
  title={Design and validation of an unmanned surface vehicle simulation model},
  author={Heins, Peter H and Jones, Bryn Ll and Taunton, Dominic J},
  journal={Applied Mathematical Modelling},
  volume={48},
  pages={749--774},
  year={2017},
  publisher={Elsevier}
}

@article{hasan2023predictive,
  title={Predictive digital twins for autonomous surface vessels},
  author={Hasan, Agus and Widyotriatmo, Augie and Fagerhaug, Eirik and Osen, Ottar},
  journal={Ocean engineering},
  volume={288},
  pages={116046},
  year={2023},
  publisher={Elsevier}
}

@article{chan2025investigating,
  title={Investigating the impact of seafarer training in the autonomous shipping era},
  author={Chan, Jevon P and Pazouki, Kayvan and Norman, Rose and Golightly, David},
  journal ={Journal of Marine Science and Engineering},
  volume={13},
  number={4},
  pages={818},
  year ={2025},
  publisher={MDPI}
}

@article{cai2013evaluating,
  title={Evaluating of marine traffic simulation system through imazu problem},
  author={Cai, Yuanbing and Hasegawa, Kazuhiko},
  journal={Proc Jpn Soc Nav Archit Ocean Eng},
  volume={17},
  pages={191--194},
  year={2013}
}

@book{Skolnik,
  author    = {Merrill I. Skolnik},
  title     = {Introduction to Radar Systems},
  edition   = {3rd},
  publisher = {McGraw-Hill},
  year      = {2001}
}

@book{Curry,
  author    = {G. Richard Curry},
  title     = {Radar System Performance Modeling},
  publisher = {Artech House},
  year      = {2005},
  isbn      = {978-1-58053-816-9}
}

@techreport{ITU838,
  title = {Specific Attenuation Model for Rain for Use in Prediction Methods},
  institution = {International Telecommunication Union},
  number = {ITU-R P.838-3},
  year = {2005}
}

@techreport{ITU840,
  title = {Attenuation Due to Clouds and Fog},
  institution = {International Telecommunication Union},
  number = {ITU-R P.840-8},
  year = {2019}
}

\end{document}